# Investigation of a Data Split Strategy Involving the Time Axis in Adverse Event Prediction Using Machine Learning

*Katsuhisa Morita*[*, 1]*, Tadahaya Mizuno*[*, †, 1]*, and Hiroyuki Kusuhara*[‡, 1]

**AUTHOR ADDRESS**

[1]Graduate School of Pharmaceutical Sciences, the University of Tokyo, Bunkyo-ku, Tokyo, 113-0033, Japan

**ABSTRACT**

Adverse events are a serious issue in drug development and many prediction methods using machine learning have been developed. The random split cross-validation is the de facto standard for model building and evaluation in machine learning, but care should be taken in adverse event prediction because this approach does not match to the real-world situation. The time split, which uses the time axis, is considered suitable for real-world prediction. However, the differences in model performance obtained using the time and random splits are not clear due to the lack of the comparable studies. To understand the differences, we compared the model performance between the time and random splits using nine types of compound information as input, eight adverse events as targets, and six machine learning algorithms. The random split showed higher area under the curve values than did the time split for six of eight targets. The chemical spaces of the training and test datasets of the time split were similar, suggesting that the concept of applicability domain is insufficient to explain the differences derived from the splitting. The area under the curve differences were smaller for the protein interaction than for the other datasets. Subsequent detailed analyses suggested the danger of confounding in the use of knowledge-based information in the time split. These findings indicate the importance of understanding the differences between the time and random splits in adverse event prediction and strongly suggest that appropriate use of the splitting strategies and interpretation of results are necessary for the real-world prediction of adverse events. We provide analysis code and datasets used in the present study (https://github.com/mizuno-group/AE_prediction).

**INTRODUCTION**

Adverse events often lead to the withdrawal of drugs during their development or after their marketing[1,2]. Accurate prediction of the potential risk of the adverse effects makes it possible to avoid the withdrawal by identifying the potential risks in advance[3,4]. Therefore, various predictive evaluation methods, such as toxicogenomics represented by TG-GATEs and DrugMatrix and high-throughput assays represented by Tox21 and ToxCast[5–7], have been developed. Among them, in silico approaches utilizing machine learning (ML) are widely used in adverse event prediction because they do not require drug synthesis or complex experiments[3,4].

For the evaluation of the prediction accuracy of ML models, most studies have adopted the random split cross-validation method[8,9]. The applicability domain (AD), which defines and restricts the applicable compounds for a model considering the chemical spaces of the training set, determines whether such a model can be applied to the test set[10]. In contrast to random split, time split is a data split method that considers when drugs were developed and prepares the training subset and the test subset based on a time axis. This split method is important in terms of real-world application since we would want to create a model that could predict adverse events of the upcoming drugs based on the existing drugs.

Li et al. developed a model for predicting drug-induced liver injury using time splits and evaluated its predictive performance[11]. The area under the receiver operating characteristic curve (AUC) value in their study was lower than those in other reports about adverse event prediction using the random split[12,13]. Although they do not strongly advocate the importance of time split in their paper, their results suggest that the accuracy of the random-split-derived model is overestimated compared to the accuracy of the time-split-derived model. Because their study focused specifically on drug-induced liver injury, used only structure-based features as input, and did not compare to random split under the same settings, it is unclear whether the random split-derived model is overestimated or whether the same claim can be made for other data sets and adverse events.

To build an adverse event prediction model, it is important to select the following three methods: input (how to describe the compound information), algorithm (how to predict the event), and evaluation (how to evaluate the performance)[14,15]. Large-scale surveys exist for both inputs and algorithms[16–19]. For example, MoleculeNet provides a comprehensive evaluation of the datasets and ML methods for many targets, and also provides the well-formed benchmark datasets[20]. However, the evaluation methods regarding data split strategy have not been comprehensively investigated in adverse event prediction and it is still unclear what to watch out for in time split. In other words, since the lack of studies comparing data split methods is an issue in this field, we examined the differences in characteristics between time split and random split in adverse event prediction in terms of prediction methods, method inputs, and target adverse events.

We prepared the comprehensive inputs and ML methods and used them to predict eight adverse events obtained from the Side Effect Resource (SIDER) database and discussed them based on the difference in the prediction accuracy of the split methods[21]. We also investigated the bias with respect to the protein interaction dataset that exhibited heterogeneity.

## MATERIALS AND METHODS

### Data Preparation

All compound names and the earliest marketing start dates were obtained from the DrugBank database[22]. To avoid shaky notation, lists of synonyms were extracted from the PubChem database using the PubChemPy package (ver1.0.4). Thereafter, all compound names in all datasets were standardized to these synonyms and the corresponding DrugBank notations. Compounds without a corresponding name were excluded from the analysis. We used compounds commonly present in all datasets for further analysis. The number of compounds that are common to all datasets is 451. The version information and download dates of all the datasets used in the present study are listed in Table 1. The preparation of the datasets is described in the following sections and the overall flow is shown in Figure S1.

Below are the details of each dataset.

1. **Prediction Target Dataset: SIDER**

This database contains information on marketed medicines and their adverse reactions[21]. MoleculeNet is large benchmark dataset for molecular machine learnings [20]. It organizes several public datasets and provides the open-source implementations of learning algorithms and the processed datasets. The SIDER dataset utilized in this study is derived from the processed datasets in MoleculeNet which is not the raw data available on the original SIDER website. To avoid highly imbalanced targets, the adverse events with a positive ratio of less than 0.2 or more than 0.8 in all compounds were excluded (Figure S2).

2. **Protein Interaction Dataset: DrugBank/CTD/SemMed**

These datasets are about protein interaction information. These are table data where 1 is assigned if there is a relationship between the compound and target protein, and 0 if there is not. The DrugBank or Comparative Toxicogenomics Database (CTD) dataset was downloaded from the corresponding public database. The Semantic MEDLINE (SemMed) predication data were downloaded from the public database, as indicated in Table 1. Because this dataset is an aggregation of relationships between subjects, predicates, and objects, interaction information was converted into a dummy variable. In brief, if there was even one sentence in the dataset where some specific predicate existed between a target compound and a target gene, it was judged that a relationship existed and a value of 1 was assigned. Because of the large number of features, only the proteins that were common to these three datasets were used as the features for interpretability. In preliminary studies, this intersection procedure did not lead to a significant difference in the prediction accuracy (Figure S3). As for SemMed and CTD datasets, we have not received permission to redistribute them, these datasets are not provided in sample code.

3. **Transcriptome Profile Dataset: L1000**

The L1000 dataset is derived from a transcriptome profile database, which collected transcriptome data of cells treated with lots of low molecular weight compounds. This dataset is data-driven and available without existing knowledge. This dataset was downloaded from the iLINCS database[23]. The "Value_LogDiffExp" values were extracted and used as the gene expression change data.

4. **Structure-based Dataset: Mold2/Mordred/Mol2vec**

These datasets are about structure-based information. Structure information can be converted into numerical features using each algorithm. Two datasets (Mold2/Mordred) are created with rule-based algorithms and therefore have no potential for data leakage. Mol2vec data set has the potential for data leakage depending on when it was created because it is a machine learning based method that learns existing knowledge. All compounds were converted to the canonical simplified molecular input-line entry system format using the PubChemPy package (ver1.0.4). Then, the SMILES strings were converted to molecular information using the RDKit package (ver2021.03.5). Molecular information was subjected to each conversion Python package or software.

5. **Chemical Property Dataset: PubChem**

PubChem database contains computed chemical properties such as molecular weight, XLogP. Chemical property information was extracted from PubChem database using the PubChemPy package.

6. **ADMET Dataset: ADMET Predictor**

ADMET Predictor software (version 10.2.0.15; Simulations Plus, Inc., CA, USA) is a machine learning platform for predicting ADMET parameters. This software was used to analyze the molecular information of all compounds. Features expressed as numerical values were used as is, while the rest were converted into dummy variables. This dataset may be suffered from data leakage depending on when the model was created.

**Feature Selection**

We adopted a filtering method for each dataset. First, the duplicated features were excluded. Second, features that had a coefficient of variation value < 0.05 were excluded. Finally, one of each pair of features with a high pairwise correlation (r2 > 0.85) was dropped.

**Data Split**

In the time split, all compounds were sorted by the marketing start date obtained from the DrugBank database and partitioned into training and test compounds. The date 1998/10 was used as a threshold because the ratio of the training compounds (361) to test compounds (90) became approximately 4:1. In the random split, all compounds were randomly partitioned into 361 training compounds and 90 test compounds 20 times without any stratification of the positive and negative ratios.

**Model Building**

Each training dataset was randomly split into five equal parts with the same positive and negative ratios. Four parts were used as training datasets while the remaining part was used as a validation dataset. This training was repeated five times for different part combinations, and five models were generated. Prediction probability was calculated using an average of predictions of the five models. Random states were changed when creating the five models inside the training data (splitting seed and training seed) and a series of model building was repeated for 20 times. A flow chart of model building is presented in Figure S4.

**ML Methods**

We adopted six popular ML methods, as shown in Table 2. All ML algorithms were executed in Python (version 3.9.6). The version information and package names of all algorithms are also listed in Table 2. The hyperparameters of each ML model were tuned for all concatenated dataset using the out of fold AUC as an index. The same parameters were used for the analysis of each dataset, as preliminary studies confirmed that the accuracy did not vary significantly depending on the parameters. All used hyperparameters are uploaded in the github (https://github.com/mizuno-group/AE_prediction) and Supplementary Data 6.

**Performance Evaluation**

The following four indicators were used to assess the prediction performance: accuracy, F1 score, Matthews correlation coefficient, AUC, and area under the precision-recall curve. Detailed results can be found in the Supplementary Data 1 and 5.

**Feature Importance**

Feature importance was calculated using the permutation importance method. Each training dataset was randomly split into four equal parts with the same positive and negative ratios. Three parts were used as training datasets while the remaining part was used as a validation dataset. This training was repeated four times for different part combinations, and four models were generated. For the feature k, a single column of feature k in the test dataset was randomly reordered and the AUC value was recalculated. This reordering was repeated 25 times using different random state, and an average of the AUC values was defined as $AUC_k$. Furthermore, $AUC_{all}$ was defined as the AUC value obtained using all features without any reordering. Then, feature importance was calculated as the decrease in the AUC value by the following equation.

$$Feature\ Importance = AUC_{all} - AUC_k$$

**Visualization of Chemical Space**

The difference between the distribution of the training and test data in the chemical space was compared using several methods as follows.

1. **Dimension Reduction: UMAP/PCA/t-SNE**

   All concatenated data that had the filtered features were standardized to z scores, and NaN values of the data were replaced with zero. Then, dimensional reduction of the data was performed using Uniform Manifold Approximation and Projection (UMAP), principal component analysis (PCA), or t-distributed stochastic neighbor embedding (t-SNE).

2. Structural Similarity

Molecular information was converted into extended-connectivity fingerprints (ECFPs) using the RDKit package. The similarity of structure between compounds was calculated using the Tanimoto coefficients of ECFPs.

3. Network Shortest Path Length

This section describes a method to create network of compounds based on similarity of features and to calculate the distance between compounds in terms of path length on the network. All concatenated data that had the filtered features were standardized to z scores, and NaN values of the data were replaced with zero. The adjacent matrix was calculated as correlation between compounds. In order to create complex networks from the adjacency matrix, we have applied planar maximally filtered graph algorithm, which filters out complex networks by keeping only the main representative links based on planarity[24]. All pairs of the shortest path lengths in the pruned network were computed using the NetworkX package (ver2.6.3).

**Statistical Analysis**

Statistical analysis was performed by paired samples one-sided t-test. In some results, p-values were combined using Stouffer Method to grasp the global trend and presented as a combined p-value. Data were analyzed using the scipy library of Python 3.

## RESULTS

### Comparison of Prediction Scores between Time and Random Splits

Nine various input datasets and six ML methods were used to predict eight adverse events. All compounds were divided into the training and test compound sets using the time or random split, and prediction performances were evaluated. To compare the differences in prediction scores derived from the two splitting methods, we evaluated differences in several measures of prediction, such as those in the AUC values between the time and random splits for each dataset and each ML method. The results are presented in a box plot for each adverse event in Figure 1A. Overall, random split tended to show higher AUC values than did time split, except for metabolism and nutrition disorders and blood and lymphatic system disorders (combined p-value=2.03e-23). This trend was confirmed by other calculated indicators in Figure S5 and statistical tests in Supplementary Data 5. In particular, for hepatobiliary disorders, which had the largest differences, the median AUC difference was 0.072. This finding suggests that the adverse event prediction for compounds that were divided into the training and test sets using the time split was difficult. To exclude the possibility of artifact originating from sampling effects, we performed similar analysis using different threshold (1993/1). Then, we confirmed the same trend as described above, which also supports the trend was not due to randomness from sampling effects (Figure S6).

Next, we focused on the prediction scores in detail. Figure 1B shows the AUC value of each dataset and ML method in the prediction of hepatobiliary disorders. The AUC value for the L1000 dataset was almost close to 0.5 and did not contribute to the prediction of hepatobiliary disorders using both time and random splits. Regarding the nonprotein interaction datasets, the random split had better prediction scores regardless of the ML methods or datasets. In contrast to the trends of the other datasets, the AUC value obtained using the time split was higher than that using the random split for the SemMed dataset in the case of all ML methods. This finding suggests that the SemMed dataset had different properties in this prediction. Figure 1C shows the AUC values in the prediction of metabolism and nutrition disorders, which exhibited no differences between the time and random splits. Unlike hepatobiliary disorders, the overall AUC value for this prediction was close to 0.5, while the CTD dataset, which is one of the protein interaction datasets, had a high AUC value of approximately 0.8. This finding indicates that the CTD dataset had different properties in this prediction.

The distinctiveness of protein interaction datasets motivated us to reanalyze our first result by dividing the datasets into protein interaction datasets (DrugBank/CTD/SemMed) or "other" datasets. It was confirmed that the difference between the time and random splits tended to be smaller for the protein interaction datasets than for the other datasets, except for reproductive system and breast disorders and blood and lymphatic system disorders (Figure 1D). These findings indicate that the protein interaction datasets had some different characteristics than did the other datasets in adverse event prediction using the time split.

**(A)** Combined p-value=2.03e-23

**(B)** Hepatobiliary disorders

**(C)** Metabolism and nutrition disorders

**(D)**
- Hepatobiliary disorders ***
- Metabolism and nutrition disorders **
- Eye disorders ***
- Reproductive system and breast disorders p=0.0909
- Endocrine disorders ***
- Blood and lymphatic system disorders p=0.363
- Renal and urinary disorders p=0.0740
- Ear and labyrinth disorders ***

Combined p-value=4.75e-14

**Figure 1.** Comparison of prediction scores between time and random splits. Six machine learning methods, nine input datasets, and the time and random splits were used to predict eight adverse events and evaluate prediction accuracy. (A) Differences in the AUC values between the time and random splits were calculated for all patterns, summarized for each adverse event, and visualized using a box plot. The AUC values for (B) hepatobiliary disorders or (C) metabolism and nutrition disorders were visualized using heat maps. Note that the lower and upper limits of the heat maps are 0.5 and 0.8, respectively. (D) Differences in the AUC values between the time and random splits for all patterns were divided into the protein interaction datasets (DrugBank/CTD/SemMed) and other datasets, and further visualized using box plots for each adverse event. All tests of significance were conducted using a paired samples one-sided t test: *p < 0.05, **p < 0.01, ***p < 0.001. The combined p-value was calculated by integrating p-values using the Stouffer Method.

**Comparison of Chemical Spaces in the Time Split Datasets**

Previous studies suggested that AD should be considered in the application of ML methods for toxicity prediction[25,26]. We compared the distributions of the training and test compounds when divided using the time split by visualizing them in several ways.

Dimension reduction of all concatenated data was performed using the following three methods: UMAP, PCA, and t-SNE. A scatterplot of the reduced components of the training and test compounds showed that the test compounds tended to be close to each other; however, no large difference was found between the training and test compounds (Figure 2A; Supplementary Figs. 7A and 7B). These tendencies were also confirmed using other calculation methods for chemical space distances, including calculation of the cosine distances of all concatenated data (Figure 2B); structural similarity of ECFPs, which is one of the major fingerprint methods (Figure 2C); and the shortest path length of the compound network created using the planar maximally filtered graph algorithm (Figure 2D). These findings suggest that the chemical spaces of the training and test compounds using the time split were similar and insufficient for deciding which model should be used.

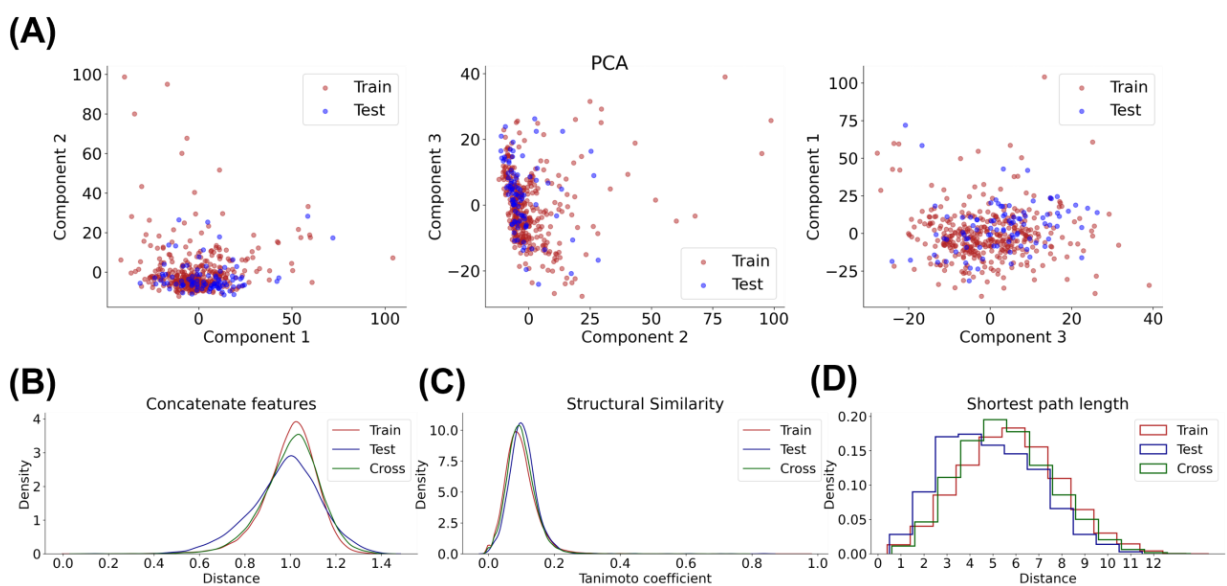

**Figure 2.** Visualization of chemical spaces. The chemical spaces of the compounds used in this study were visualized using various methods. We merged all the datasets used in this study and conducted feature reduction using a filtering method (all data). (A) Dimensionally-reduced data obtained using principal component analysis of all data were plotted. (B) Correlation distances between all data points within the training compounds, the test compounds, and between the training and test compounds were plotted on kernel density estimate plots. (C) The Tanimoto coefficients for ECFPs of the training compounds, the test compounds, and between the training and test compounds were plotted on kernel density estimate plots. (D) We created a compound network from all data using the planar maximally filtered graph algorithm; calculated the shortest path length within the training compounds, the test compounds, and between the training and test compounds; and visualized them using a histogram.

**Investigation of the SemMed Dataset in the Prediction of Hepatobiliary Disorders**

Next, we decided to investigate in detail the different properties of the protein interaction datasets shown in Figure 1. We focused on hepatobiliary disorders because of their higher prediction scores and larger differences in the AUC values between the time and random splits compared with the other organ targets. To investigate the features that contributed to the prediction using the time split, the feature importance was calculated using the permutation importance method and XGBoost[27]. The features were sorted in descending order, and the top 20 were plotted as a bar graph (Figure 3A). We found that the top five proteins/features had been previously reported to be involved in the injuries of the liver, which was the target organ of this prediction[28–32]. SemMed is a database that extracts semantic relationships from article titles and abstracts in PubMed. Such binding information may have been collected after the occurrence of liver injuries to investigate the mechanism of the toxicity. To investigate this possibility, we compared the time lag between the Food and Drug Administration (FDA) approval date for the compound and the publication date of the first article reporting the interaction information for the target protein. Figure 3B shows the histograms of time lags for all pairs of the test compounds and proteins and each time lag for the top five proteins. In the test compounds, the publication dates of articles about proteins that had higher feature importance tended to be earlier than those of articles about proteins that had relatively lower feature importance, except for the protein ABCB1. This trend was statistically confirmed using a permutation test for the top 15 proteins (Figure 3C). These findings support the possibility of data leakage of the knowledge-based protein interaction.

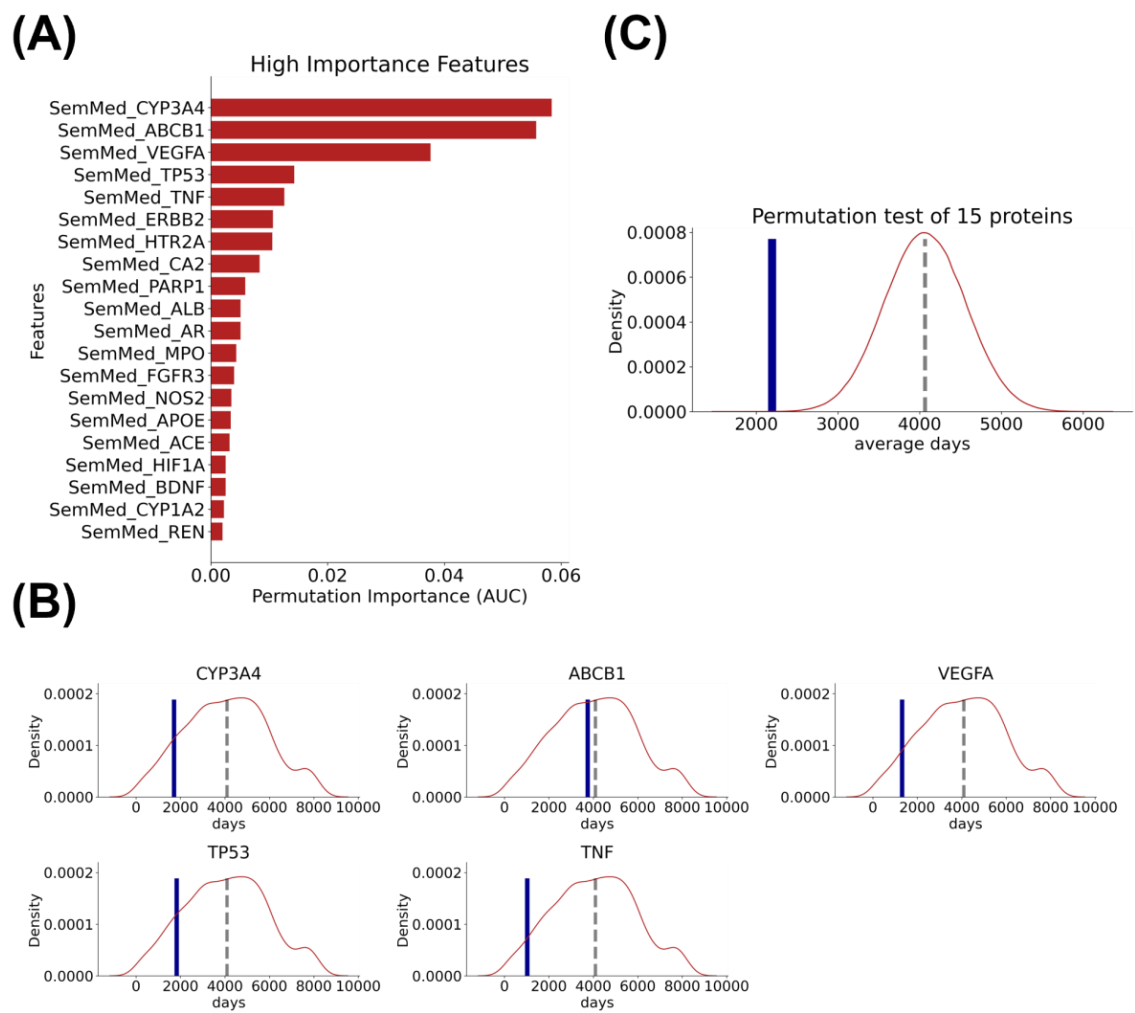

**Figure 3.** Important features of the SemMed dataset in the prediction of hepatobiliary disorders. (A) Feature importance of the SemMed dataset in the prediction of hepatobiliary disorders using XGBoost and the time split was calculated using the permutation importance method and the top 20 features are shown. (B) For the top five features, the difference between the FDA approval date and the publication date of the first article (time lag) in the test compounds is displayed as a blue bar. The time lags for all the features are shown as a kernel density estimate plot (red curve). The median value of the time lags is plotted as a gray dashed line. (C) The average time lags of the randomly selected 15 features were calculated using 100,000 times permutation (red curve). The average time lag of the top 15 features is shown as a blue bar.

**Real-world Prediction of Hepatobiliary Disorders**

Finally, we evaluated in detail the prediction of hepatobiliary disorders using the time split and the appropriate inputs for real-world prediction based on the above findings. The protein interaction, ADMET, and L1000 datasets were excluded because the two former datasets had the potential for data leakage and the latter did not contribute to the prediction (Figure 1B). We evaluated the prediction scores using the following two models based on the use of features: a concatenation model and an ensemble model. The features of all datasets were simply concatenated in the former model, which is widely used in the field of toxicity or property prediction. In the latter approach, a model is constructed for each dataset and the average of all the probabilities is used as the output. As shown in Figure 4A, the superiority of the concatenation or ensemble model with regard to the prediction performance differed depending on the splitting and ML methods. The prediction with the highest AUC value in the time split was obtained using XGBoost on the concatenated datasets. This AUC value was approximately 0.68, which is considered low in the field of ML.

To investigate which features contributed to the prediction with the highest AUC value in the time split, the feature importance was calculated using the permutation importance method. The top 20 features that had higher feature importance are shown in Figure 4B. Most of the high-importance features were obtained from the structures of the compounds. We visualized the seven compounds for each of the top (high) and bottom (low) features of ATSC5i and BCUTd-1l in all compounds. The compound structure is labeled using the compound name, the direction of the feature value (high or low value), and the label of hepatobiliary disorders (1 in the compound name indicates that "hepatobiliary disorders has been reported"). The visualized compound groups showed a large number of functional chemical groups, such as the hydroxy groups, for the first important feature, ATSC5i, and nitrogen atoms with triple bonds for the second important feature, BCUTd-1l.

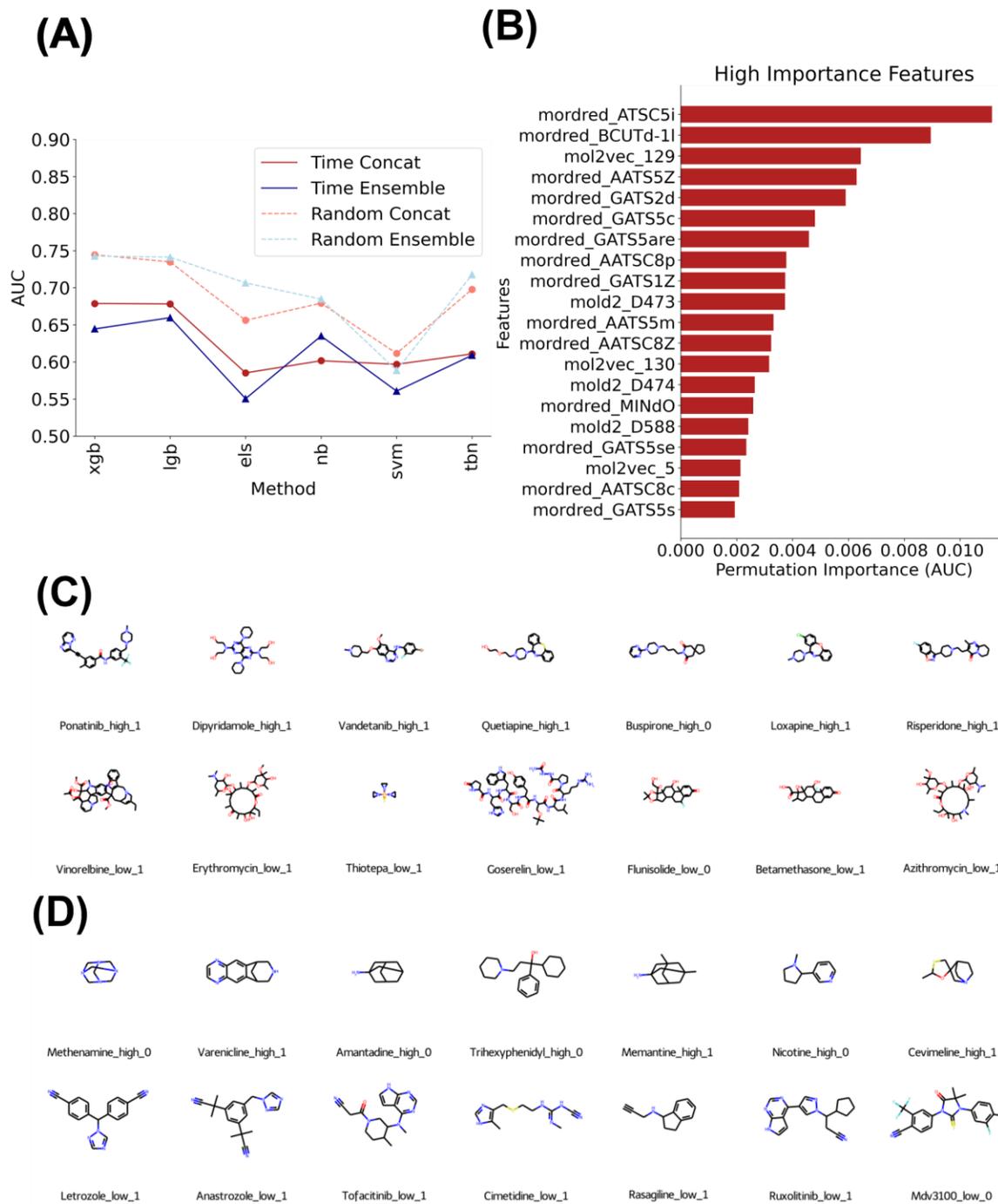

**Figure 4.** Prediction accuracy for hepatobiliary disorders and important features. Five datasets and two split methods (the time and random splits) were used to evaluate the prediction accuracy for hepatobiliary disorders. (A) The AUC values for the concatenation and ensemble models are shown. (B) The feature importance was calculated using the permutation importance method to investigate which features contributed to the prediction with the highest AUC value in time split, which was obtained using XGBoost on the concatenated datasets. The top 20 features are shown. Seven compounds for each of the top (high) and bottom (low) features of (C) ATSC5i and (D) BCUTd-1l in all compounds were visualized. The compound structure is labeled using the compound name, the direction of the feature value (high or low value), and the label of hepatobiliary disorders (1 in the compound name indicates that hepatobiliary disorders has been reported).

## DISCUSSION

### Importance of Adopting Time Split Instead of Random Split

In the field of ML, which deals with datasets where the time axis is important, such as temperature or stock price datasets, model building or data split takes time series into account. Typical examples of the latter are the time-series split cross-validation method, in which the split points are set in a time-series order, and a method in which the test samples are predicted one-by-one while updating the model for each prediction. Although this approach could be applied to the present dataset, we employed a split at a single time point in the present study for comparison with the commonly used random split cross-validation method. For many targets, the prediction accuracy was rather low, especially using the time split, although simple models were used for the comparison to ensure easy interpretability (Figure S8). In addition, we compared the prediction accuracy of the random split with the same positive ratio as that of the test split, but the difference in the AUC values still existed. This finding suggests that the difference in the positive ratios between the time and random splits does not explain the difference in prediction accuracy (Figure S9). One of the possible explanations for the accuracy in the time split being lower than that in the random split is the difficulty to evaluate novel drug groups accurately. For instance, the test compounds in the time split included rapamycin and temsirolimus, which have similar characteristic structures, and it was confirmed that predictions for both of these drugs failed during the time split. The groups of compounds that have similar pharmacological effects or characteristic structures generally tend to have similar adverse events[33,34]. Thus, the models using the random split can capture the tendency and features of compound classes and explain the superiority of prediction accuracy in this splitting method. However, the random split may not be suitable for real-world prediction, particularly in cases of novel modality. Thus, the use of the time split should be considered in adverse event prediction.

**AD Considerations**

The concept of AD is essential for prediction in general[35]. In the field of quantitative structure-activity relationships, the importance of the AD has been emphasized in terms of real-world prediction[26,36]. The AD in the field of quantitative structure-activity relationships is mainly evaluated and discussed based on its distribution in the chemical space based on fingerprints. Considering the possibility that the poor accuracy of the time split was caused by inappropriate applicability, we investigated the AD. In fact, it is clear that there are some biases in the test dataset, such as anti-human immunodeficiency virus drugs, angiotensin converting enzyme inhibitors, and tyrosine kinase inhibitors (compound list available on the github page). However, the distribution of chemical spaces did not show any significant difference between the training and test compounds even though several dimension reduction methods were tested (Figure 2A; Supplementary Figs. 4A and 4B). These findings suggest that a decision of model application based on the AD would be misleading, in particular when the AD is defined using the distance-based approaches involving simple fingerprints, such as ECFPs, although they are widely used. Thus, care should be taken in the use of the AD based on distances, and advances in describing the properties of compounds are expected to improve this situation. Moreover, other approaches for defining the AD, such as prediction-based methods[37,38], are actively studied. The progress of such methods for defining the AD is expected to improve prediction performance by refining the process of model application.

**Heterogeneity of the SemMed Dataset in the Prediction of Hepatobiliary Disorders**

Next, we focused on the reporting bias of the protein interaction datasets. The structural information of a compound is constant and for the present dataset, it was uniquely determined regardless of the acquisition dates. Furthermore, protein interaction information is knowledge-based and time-dependent, which means that there is a reporting bias and a possibility of data leakage in the studies for adverse event prediction. Thus, we investigated the possibility of data leakage by comparing the time lag between the FDA approval date and the publication date using the SemMed dataset. Although it would be best to use the date when the adverse event was recognized, the FDA approval date was used instead because of the difficulty in data collection. Notably, the important features (proteins that interacted with a compound of interest) in the prediction using the time split were reported after the FDA approval statistically earlier than the other proteins (Figure 3C). One possible reason for this finding could be that the interaction information was reported as a part of a study to determine the cause after the side effect had been reported. Such a scenario would lead to a strong bias and data leakage. Moreover, the top five proteins of importance in the prediction of hepatobiliary disorders using the SemMed dataset were CYP3A4, ABCB1, VEGFA, TP53, and TNF, which were previously reported to be associated with liver injury[28–32]. In addition, the number of reports regarding the proteins that play essential roles in xenobiotic metabolism, such as CYP3A4 and P-glycoprotein, is highly dependent on the timing of the guidelines issued by each regulatory agency (e.g., FDA, Pharmaceuticals and Medical Devices Agency). Thus, although protein interaction information is useful in toxicity prediction because of its biological relevance, care should be taken when we taking into account the such time-axis.

**Limitations of the Current Prediction**

Finally, we investigated how much prediction accuracy can be achieved using models based on the datasets without protein interaction information, to exclude the time-dependent bias for real-world prediction. The difference in the prediction accuracy between the time and random splits continued to exist even after combining the five different structural information datasets. Notably, the ensemble model showed relatively good performance. This finding indicates that the ensemble approach for datasets can be used in the time split as well as random split evaluation, although the prediction accuracy depends on the ML methods and the best score was achieved in the simple concatenation approach using XGBoost, which is a well-employed combination, in the time split.

Moreover, the highest AUC value in the time split was low and <0.7. Thus, care should be taken regarding the reliability of the findings. However, the important features in the model were consistent with those in previous reports about the characteristics of chemical structures related to adverse events, i.e., the high reactivity derived from the unstable structure of a compound can lead to toxicity[39,40] (Figs. 4C and 4D). Although only the simple concatenation or ensemble model was employed in the present study for interpretability and comparability, it is expected that the prediction accuracy of the time split could be improved using complex models as reported in previous studies[41,42]. However, how much such improvement would contribute to solving the intrinsic problem of poor performance for real-world prediction remains to be seen. In the present study, the ADMET dataset did not have much impact on prediction accuracy, but it contains essential and distinctive information that can be used for directly linking drugs and organisms based on pharmacokinetics. The effective use and integration of such information in adverse event prediction using ML would aid the real-world prediction of drug toxicity.

**CONCLUSION**

In the present study, we confirmed that the random split tended to be overoptimistic compared with the time split in various combinations of inputs, outputs, and algorithms. We also found that in adverse event prediction, a simple approach involving AD based on chemical space, which is widely used, could be misleading with regard to real-world prediction. Furthermore, care should be taken in the use of protein interaction datasets for predictions using the time split. One of the major concerns for the time split evaluation is the limited amount of available data. Although it is essentially inevitable, recent advances in the organization and use of medical big data promote data-to-knowledge conversion[43]. For real-world prediction of drug toxicity, it is important to accumulate toxicity data that are easily accessible and associated with time information, and we provide a small portion of such data used in the present study (https://github.com/mizuno-group/AE_prediction).

# TABLES

**Table1.** Information regarding the nine datasets used in the present study

| Name | No. of features | Acquisition date | Version | PMID | Reference |
|---|---|---|---|---|---|
| **DrugBank** | 478 (/ 882) | 2021/04 | 5.1.9 | 29126136 | [22] |
| **CTD** | 478 (/ 21650) | 2021/12 | 16679 | 33068428 | [44] |
| **SemMed** | 478 (/ 5155) | 2021/12 | 43 | 28678823 | [45] |
| **L1000** | 958 | 2021/06 | - | 29195078 | [23] |
| **Mold2** | 644 | 2021/06 | 2.0 | 18564836 | [46] |
| **Mol2vec** | 300 | 2021/06 | 0.1 | 29268609 | [47] |
| **Mordred** | 1517 | 2021/06 | develop branch | 29411163 | [48] |
| **PubChem** | 30 | 2021/10 | 1.0.4 | 33151290 | [49] |
| **ADMET** | 496 | 2021/08 | 10.2.0.15 | - | - |

**Table2.** Six machine learning methods used in the present study

| Method name | Abbreviation | Package name | Version |
|---|---|---|---|
| **Xgboost** | xgb | xgboost | 1.4.0 |
| **LightGBM** | lgb | lightgbm | 3.2.1 |
| **Elasticnet** | els | sklearn | 0.24.2 |
| **NaiveBayes** | nb | sklearn | 0.24.2 |
| **Support vector machine** | svm | sklearn | 0.24.2 |
| **TabNet** | tbn | torch/pytorch_tabnet | 1.9.0+cpu/3.1.1 |

# SUPPORTING INFORMATION

Supplementary Data 1, 2, 3, and 4 correspond to detailed results of the analysis (Figure 1), important descriptors in the analysis (Figure 4), detailed results of the analysis (Figure 4), and feature importance of the analysis (Figs. 3 and 4), respectively. Supplementary Data 5 is the outcome of statistical tests in Figure 1. Hyperparameters of ML methods employed in this study is summarized in Supplementary Data 6. The uploadable datasets used in this study, as well as the essential code, are available here (https://github.com/mizuno-group/AE_prediction).

## AUTHOR INFORMATION

**Corresponding Author**

† Corresponding author: Tel: +81-3-5841-4771; E-mail: tadahaya@mol.f.u-tokyo.ac.jp

‡ Corresponding author: Tel: +81-3-5841-4770; E-mail: kusuhara@mol.f.u-tokyo.ac.jp

**Author Contributions**

T.M. coordinated the project and conceived the concept. K.M. and T.M. designed the experiments, analyzed all the data, and wrote the manuscript. H.K. revised the manuscript for intellectual content. All authors approved the manuscript before submission. *These authors contributed equally.

**Funding Sources**

This study was financially supported by the Long-Range Research Initiative [17_S05-01-2] from the Japan Chemical Industry Association and a grant-in-aid from the Takeda Science Foundation.

## DECLARATION OF CONFLICTING INTERESTS

The authors declare that they have no conflict of interest.

**ACKNOWLEDGMENT**

We thank Dr. Huixiao Hong for the redistribution permission of Mold2 data and all the investigators who contributed to the creation of all datasets used in this study.
**ABBREVIATIONS**

ML, machine learning; AUC, area under the curve; XGB, XGBoost; LGB, LightGBM; TBN, TabNet; ELS, ElasticNet, SVM; support-vector machine; NB, naïve bayes.

**Supplementary Figure**

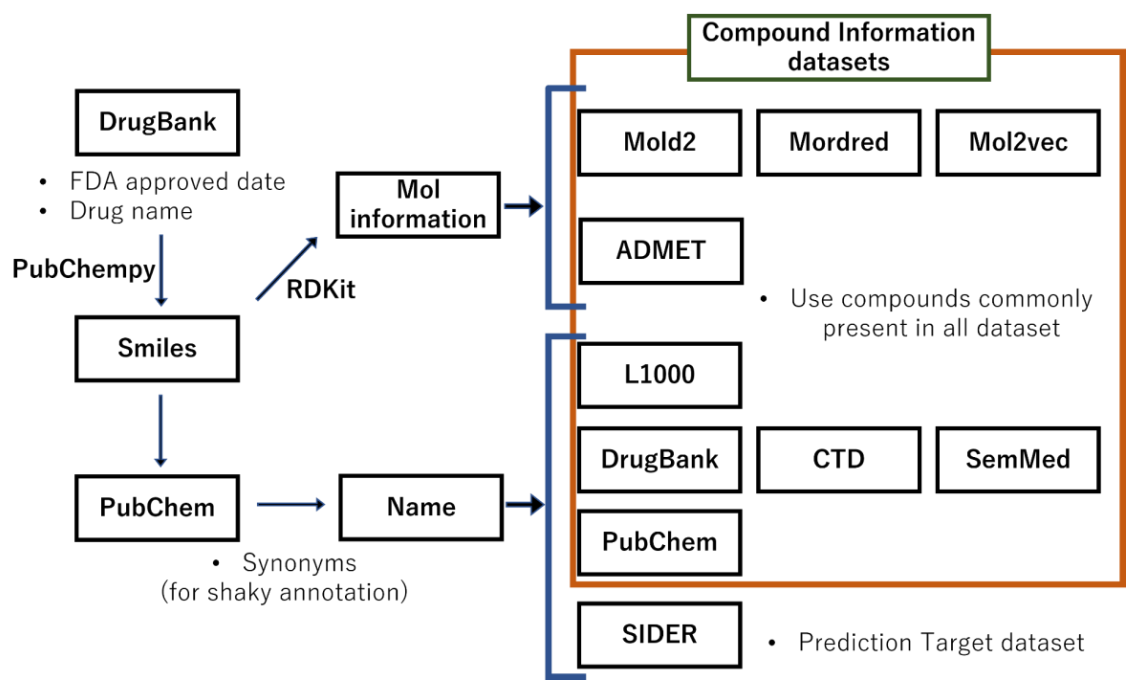

**Figure S1**

A flowchart of the data preparation in this study.

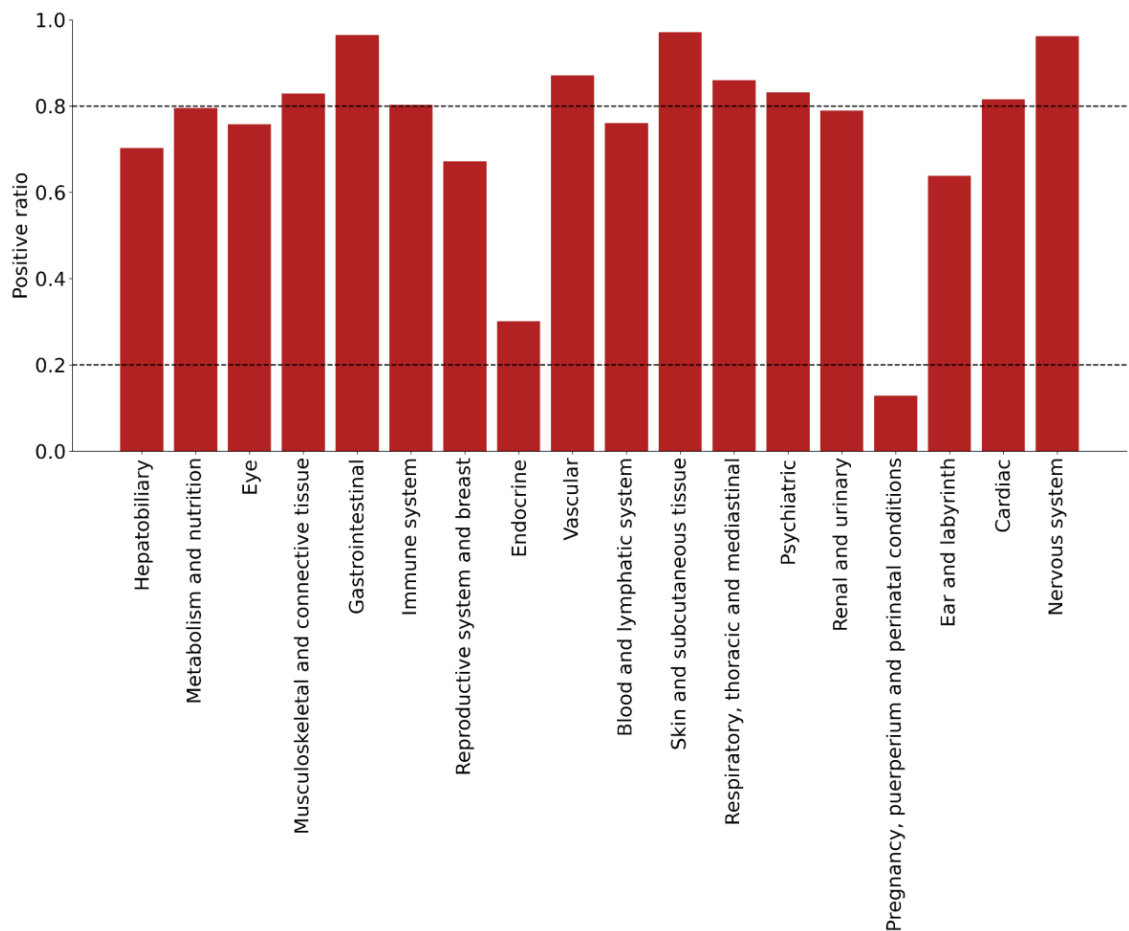

**Figure S2**

Positive ratio of all adverse events in the SIDER dataset among all compounds used in this study.

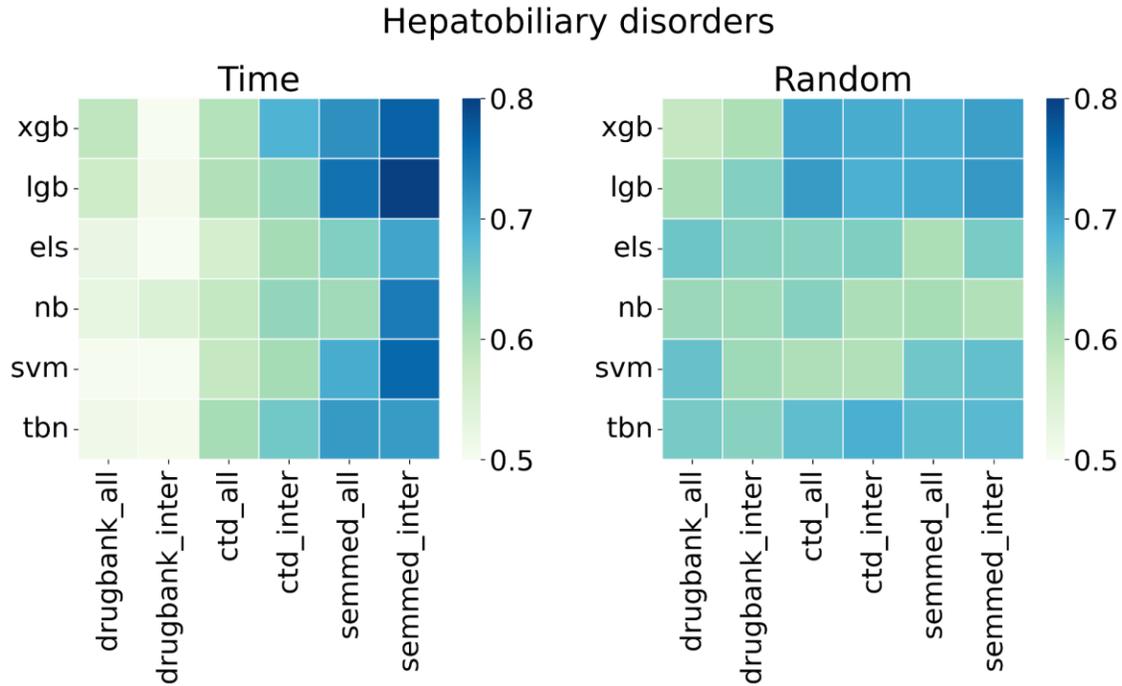

**Figure S3**

For the protein interaction datasets, the difference in area under the curve between the case using all features (DrugBank: 882, CTD: 21650, SemMed; 5155) and the case using only common protein features to predict hepatobiliary disorders was visualized by heatmap.

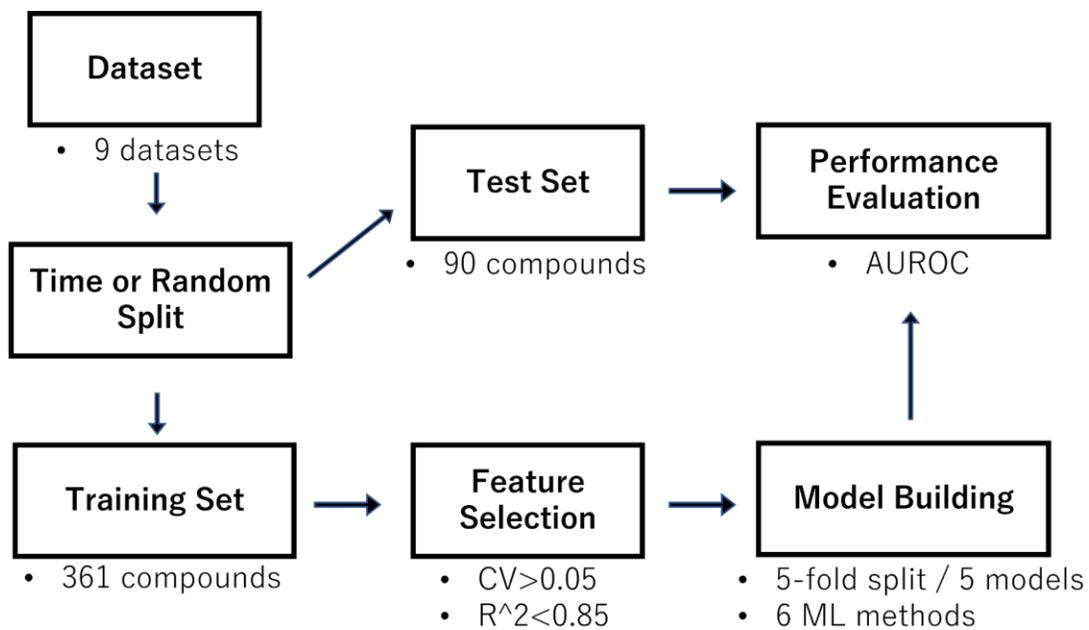

**Figure S4**

The flowchart of the model construction and the prediction performance evaluation used in this study.

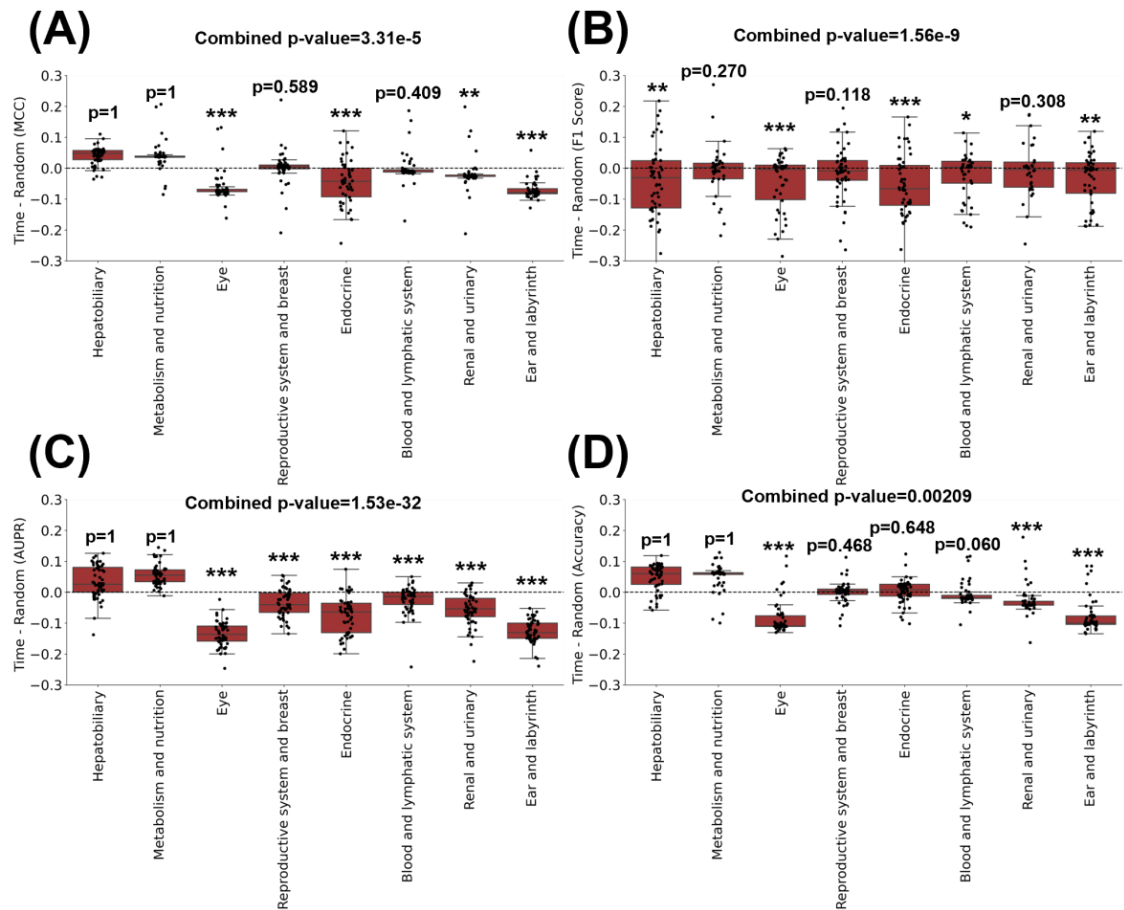

**Figure S5**

Comparison of prediction scores between time and random splits. Six machine learning methods, nine input datasets, and the time and random splits were used to predict eight adverse events and evaluate prediction accuracy. (A) Matthews correlation coefficient, (B) F1 score, (C) area under the precision-recall curve, and (D) accuracy (the same result to the Figure.1 with different evaluation metrics). All tests of significance were conducted using a paired samples one-sided t test: *p < 0.05, **p < 0.01, ***p < 0.001. The combined p-value was calculated by integrating p-values using the Stouffer Method.

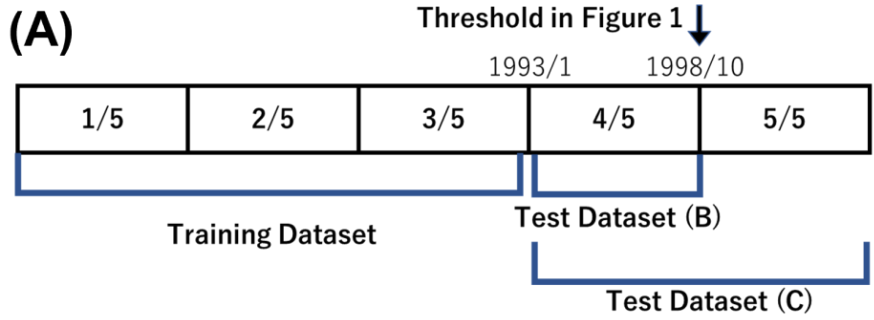
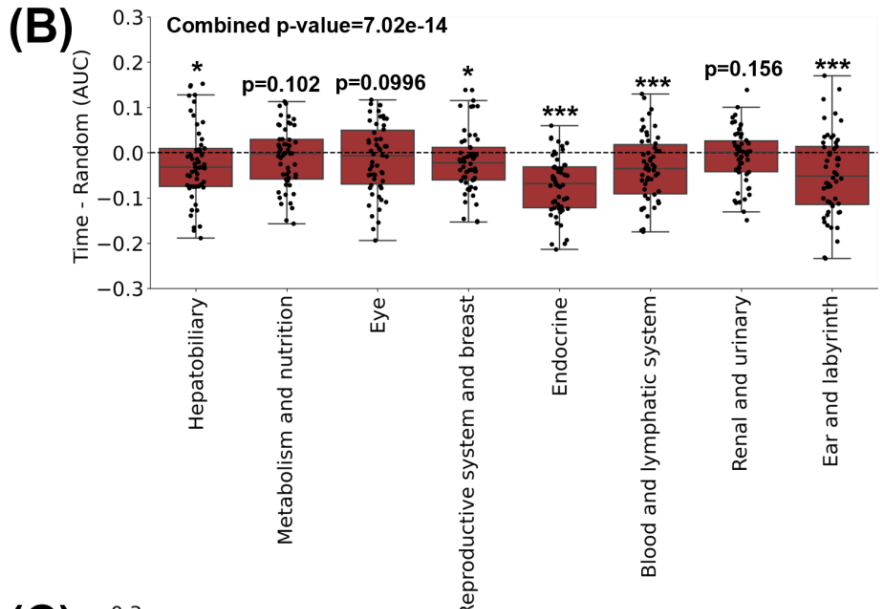
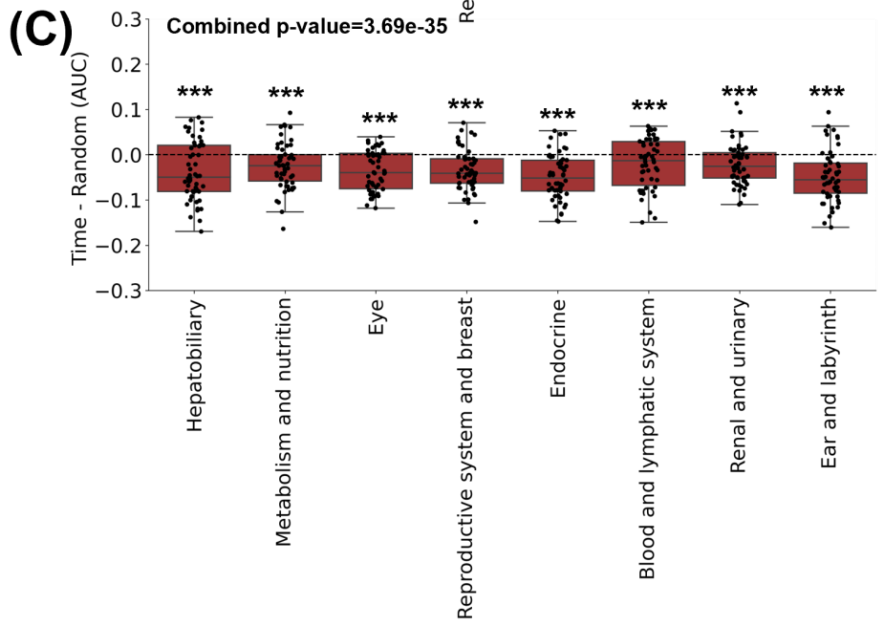

**Figure S6**

Comparison of prediction scores between time and random splits. Six machine learning methods, nine input datasets, and the time and random splits were used to predict eight adverse events and evaluate prediction accuracy with different threshold (1993/1) to the Figure, 1 analysis (1998/10). The partitioning of the training and test datasets is shown in (A). The differences area under the curve between time split and random split are visualized in (B) and (C). All tests of significance were conducted using a paired samples one-sided t test: *p < 0.05, **p < 0.01, ***p < 0.001. The combined p-value was calculated by integrating p-values using the Stouffer Method.

**(A)**

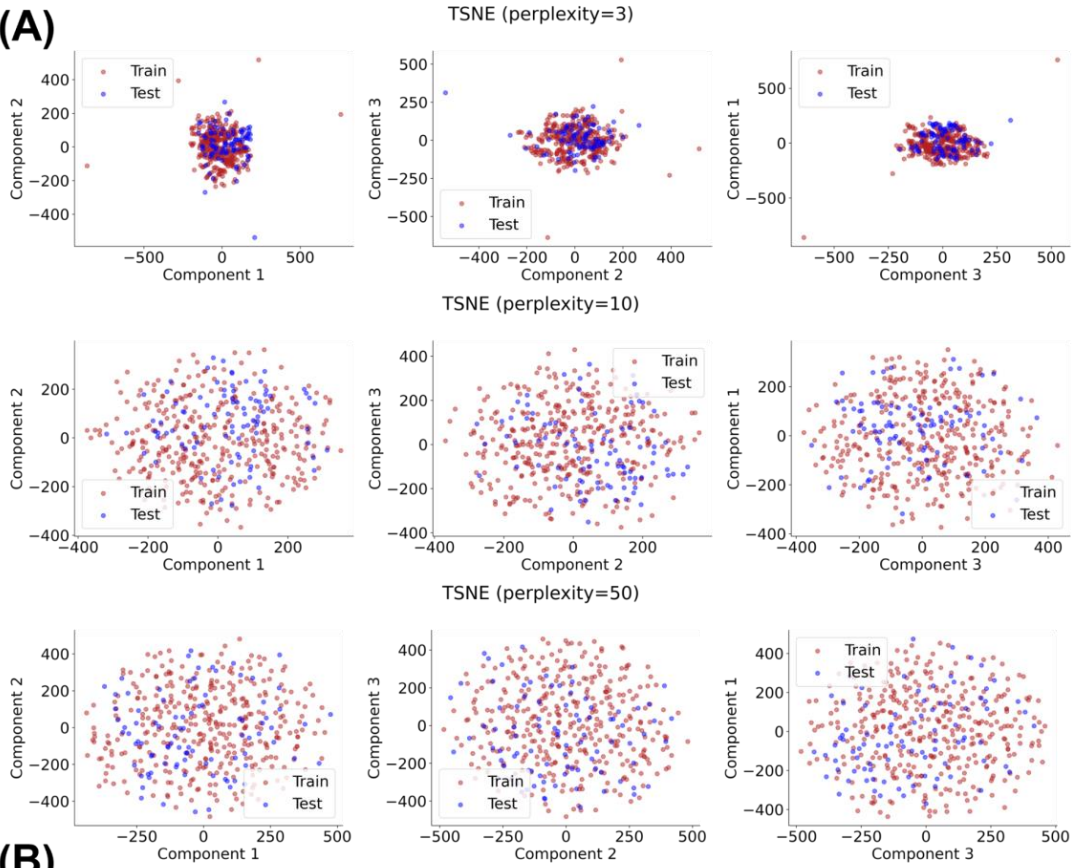

**(B)**

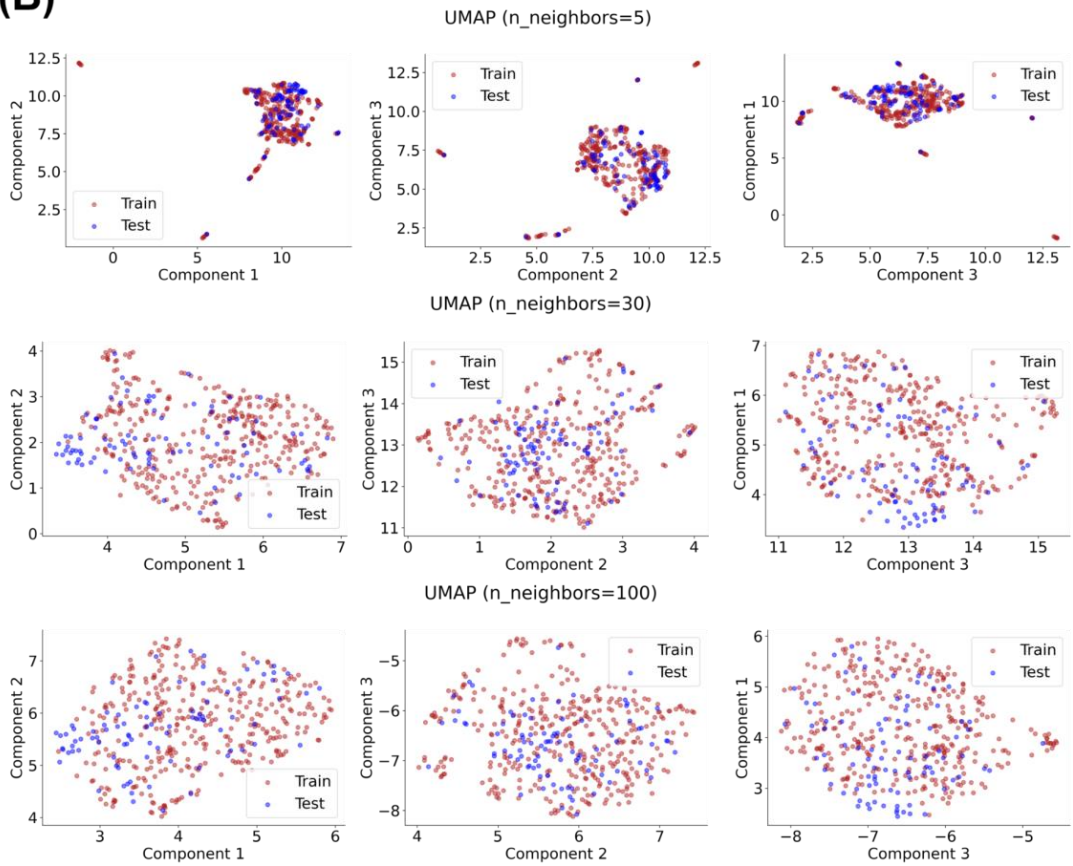

**Figure S7**

The chemical spaces of the compounds used in this study were visualized using various methods. We merged all the datasets used in this study and conducted feature reduction using filtering method (all data). Dimensionally-reduced data obtained using (A) t-SNE with 3 different hyper parameters (perplexity=3, 10, and 50) or (B) UMAP with 3 different hyper parameters (n_neighbors=5, 30, and 100) of all data were plotted.

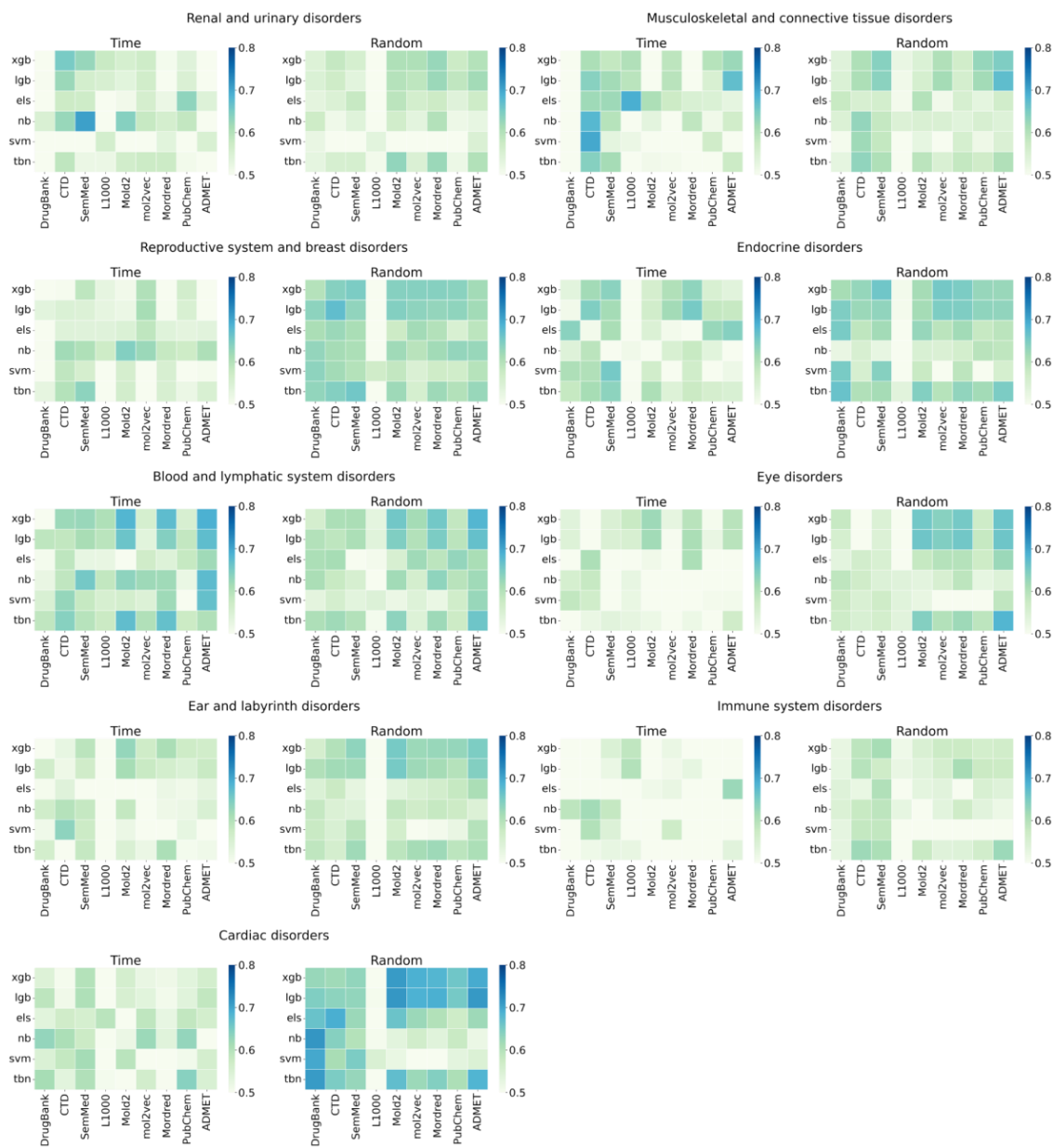

**Figure S8**

Six machine learning methods, nine input datasets, and time and random splits were used to predict nine adverse events which are not listed in the figures of the main article and evaluate accuracy. The value of area under the curve of nine adverse events were visualized using heatmap. Note that the lower and upper limits of the heatmaps are 0.5 and 0.8, respectively.

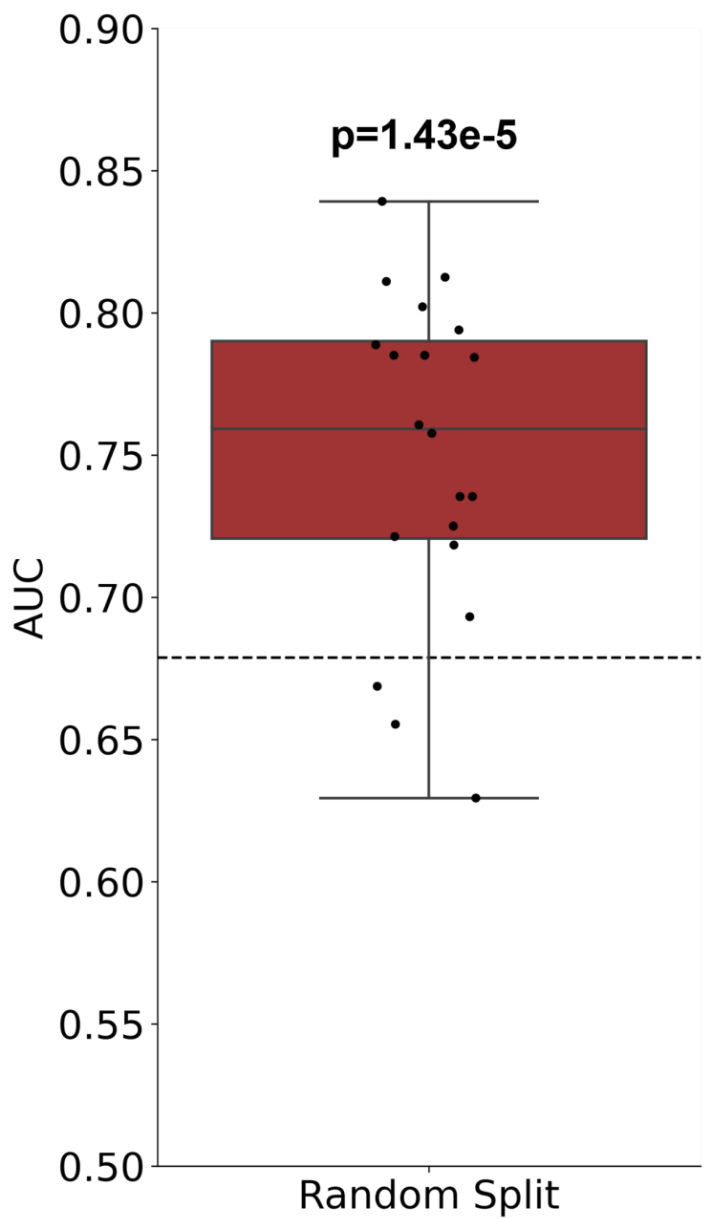

**Figure S9**

We fixed the positive ratio of the test compounds in random split as the same to that in the time split (19/90) and evaluated the prediction performance of hepatobiliary disorders 20 times with the different random state. We used the merged datasets with feature reduction using filtering method as input and XGBoost as ML method. The value of area under the curve of time and random splits are shown. The test of significance was conducted using a one-sided t test.